\documentclass[11pt]{article}
\usepackage{coling2020}
\usepackage{times}
\usepackage{url}
\usepackage{latexsym}

\usepackage{times}

\usepackage{microtype}
\usepackage{booktabs}

\usepackage{comment}
\usepackage{todonotes}
\usepackage{color,soul}
\usepackage{multirow}   
\usepackage{booktabs}
\usepackage{adjustbox}
\definecolor{mynicegreen}{RGB}{0,140,0.0}
\usepackage{longtable}
\usepackage{caption}
\usepackage{tabularx}
\usepackage{amsmath}
\usepackage{algorithm}
\usepackage{subfig}
\usepackage[noend]{algpseudocode}
\usepackage{amssymb}
\usepackage{tikz}
\usepackage{pgfplots}
\usepackage[all]{nowidow}
\makeatletter
\def\BState{\State\hskip-\ALG@thistlm}
\makeatother
\graphicspath{ {images/} }

\usepackage{enumitem}
\usepackage{import}

\usepackage[normalem]{ulem} 

\colingfinalcopy

\title{
``Judge me by my \underline{size} (noun), do you?''\\
YodaLib: A Demographic-Aware Humor Generation Framework
}

\author{Aparna Garimella$^\dagger$, Carmen Banea$^\dagger$, Nabil Hossain$^\spadesuit$ \and Rada Mihalcea$^\dagger$ \\\\
  $^\dagger$Computer Science and Engineering, University of Michigan, MI \\
  {\tt {\{gaparna,carmennb,mihalcea\}}@umich.edu}\\\\
  $^\spadesuit$Dept. Computer Science, University of Rochester, NY \\
  {\tt {nhossain@cs.rochester.edu}}}

\date{}

\begin{document}

\maketitle

\begin{abstract}
The subjective nature of humor makes computerized humor generation a  challenging task.
We propose an automatic humor generation framework for filling the blanks in Mad Libs\textsuperscript{\small{\textregistered}} stories, while accounting for the demographic backgrounds of the desired audience. 
We collect a dataset consisting of such stories, which are filled in and judged by carefully selected workers on Amazon Mechanical Turk.
We build upon the BERT platform to predict location-biased 
word fillings in incomplete sentences, and we fine-tune  BERT to classify location-specific humor in a sentence. 
We leverage these components to produce {\sc YodaLib}, a fully-automated Mad Libs style humor generation framework, which selects and ranks appropriate candidate words and sentences in order to generate a coherent and funny story tailored to certain demographics.
Our experimental results indicate that {\sc YodaLib} outperforms a previous semi-automated approach proposed for this task, while also surpassing human annotators in both qualitative and quantitative analyses.
\end{abstract}

\section{Introduction}
Computer-generated humor is an essential aspect in developing personable human-computer interactions.
However, humor is subjective and can be interpreted in different ways by different people.
Humor requires creativity, world knowledge, and cognitive mechanisms, which are extremely difficult to model theoretically.
Generating humor is considered by 
some researchers as an AI-complete problem \cite{Stock2002}.
Hence, humor generation was largely studied in specific settings; Hossain et al., \shortcite{Hossain2017} proposed a semi-automatic humor generation approach to generate humorous Mad Libs.


Language preferences vary with user demographics \cite{Tresselt64,Eckert2013,Garimella2016,Lin2018,Loveys2018}, and this has led to approaches leveraging the demographic information of users to obtain better language representations and classification performances for various NLP tasks \cite{Volkova2013,Bamman2014,Hovy2015,Garimella2017}. 
Humor is a universal phenomenon that is used across all countries, genders, and age groups \cite{Apte1985}.
Likewise, there are variations in how humor is enacted and understood due to demographic differences \cite{Kramarae1981,Duncan1990,Goodman1992,Alden1993,Hay2000,Robinson2001}.
However, to the best of our knowledge, the effect of user demographics in computational humor generation has not been studied. 

\begin{figure}[h]
    \centering
    \includegraphics[width=0.35\linewidth, height=1.75cm]{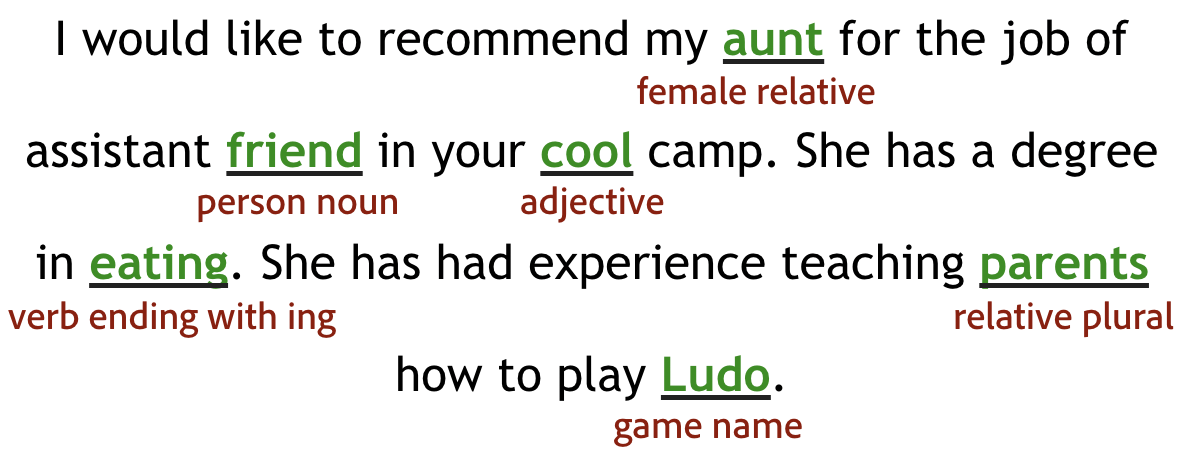}
    \caption{Example of a partial Mad Lib story.}
    \label{fig:exampleMadLib}
\end{figure}
In this paper, we introduce {\sc YodaLib}, an automatic humor generation framework
for Mad Libs\textsuperscript{\small{\textregistered}}---a story-based fill-in the blank game---which also accounts for the demographic information of the audience and story coherence. 
We use {\it location} as the demographic dimension.
{\sc YodaLib} has three stages:
(1) a candidate selection stage in which candidate words are selected to fill the sentences in each story,
(2) a candidate ranking stage to assess if a filled-in ({\it transformed}) sentence is a funny version ({\it transformation}) of the original sentence, 
and
(3) a story completion stage to join individually funny sentences to form complete Mad Lib stories that are humorous.
Fig. \ref{fig:exampleMadLib} shows an example filled-in Mad Lib.


This paper makes four main contributions:
{\bf (1)} We collect a novel dataset for location-specific humor generation in Mad Lib stories, which we carefully annotate using Amazon Mechanical Turk (AMT).\footnote{We release the location-specific Mad Lib humor dataset \url{xyz}.} 
{\bf (2)} We propose {\sc YodaLib}, a location-specific humor generation framework that builds on top of BERT-based \cite{Devlin2019} components while also accounting for location and story coherence.     {\sc YodaLib} typically generates funnier Mad Lib stories than those created by humans and a previously published semi-automatic framework.
{\bf (3)} We present qualitative and quantitative analyses to explain what makes the generated stories humorous, and how they differ from the other completions. 
{\bf (4)} Finally, we outline the similarities and differences in humor preferences between two countries: India (IN) and United States (US), in terms of certain attributes of language.
To the best of our knowledge, our work is the first computational study to automatically generate humor in a Mad Lib setting while also incorporating the demographic information of the audience, and analyzing its effect in terms of various linguistic dimensions.



\section{Related Work}
\label{sec:relatedWork}
There is a long history of research in general theories of humor \cite{Attardo1991,Wilkins2009,Attardo2010,Morreall2012,O2012,Weems2014}.
In computational linguistics, a large body of humor research involves humor recognition, and it typically focuses on specific types of jokes \cite{Mihalcea2006b,Kiddon2011,Bertero2016,Raz2012,Zhang2014,Bertero2016,Hossain2019}.
Research work on humor generation has been largely limited to specific joke types and short texts, such as riddles \cite{Binsted1997}, acronyms \cite{Stock2002}, or one-liners \cite{Petrovic2013}.
In general, it is very difficult to apply humor theories
directly to generate humor, as they require a high degree of commonsense understanding of the world. 

Owing to the subjective nature of humor, there have been recent efforts in collecting datasets for humor; Blinov et al., \shortcite{Blinov2019} collected a dataset of jokes and funny dialogues in Russian from various online resources, and complemented them carefully with unfunny texts with similar lexical properties.
They developed a fine-tuned language model for text classification with a significant gain over baseline methods.
Hasan et al., \shortcite{Hasan2019} introduced the first multimodal language (including text, visual and acoustic modalities) dataset of humor detection, and proposed a framework for understanding and modeling humor in a multimodal setting. 

In socio-linguistics, the relationship between humor and gender is widely studied.
Hay \shortcite{Hay2000} found that New Zealand women more often shared funny personal stories to create solidarity, while men used other strategies to achieve the same goal.
More recently, location has become central in sociolinguistics \cite{Johnstone2010}.
Alden et al.,~\shortcite{Alden1993} indicated that humor styles vary with countries, and humorous communications from Korea, Germany, Thailand and US, had variable content for funny advertising, while sharing certain universal cognitive structures.


The effect of demographic background on language use has gained significant attention in computational linguistics, with several efforts focused on understanding the similarities and differences in the language preferences, opinions and behaviors of people \cite{Garimella2016,Garimella2016b,Wilson2016,Lin2018,Loveys2018}.
Conversely, there has been work to leverage these demographic differences in language preferences between various groups, to develop better models for NLP tasks, such as sentiment analysis \cite{Volkova2013}, word representations \cite{Bamman2014}, sentiment, topic and author attribute classification \cite{Hovy2015}, and word associations \cite{Garimella2017}.
However, to our knowledge, none of this recent work accounts for demographic information in humor recognition or generation tasks.

We find inspiration in recent work by Hossain et al.,~\shortcite{Hossain2017}, who collected a humor generation dataset with Mad Lib stories, and proposed a semi-automated approach to aid humans in writing funny stories.
We go one step further and propose a fully-automated BERT-based demographic-aware humor generation framework. We further study the influence of location on humor preferences via AMT and seek to emulate such preferences in our automatically generated stories. 
Our work is similar to \cite{Mostafazadeh2017}, as our goal is to engage readers from different demographic groups by generating funnier versions of stories to read.

\section{Data Collection}
\label{sec:data}
Mad Libs\textsuperscript{\small{\textregistered}} is a fill-in-the-blank game \cite{price1974} to create funny stories.
A Mad Lib consists of a title and a short story, with some of the words masked. 
Players are prompted to provide replacement words for the masked entries based on the provided hints  (e.g., part-of-speech (POS) tags, {\it bodypart}, {\it food}) without having read the story. The replacement words are then filled in the story; the resulting Mad Lib is usually funny, with the humor aspect coming from the nonsensical filled-in words in an otherwise coherent and sensible story. 

We use stories curated by Hossain et al.,~\shortcite{Hossain2017}, namely {\bf Fun Libs}, as (1) Mad Libs are copyrighted and hence it is difficult to release datasets, and (2) experimentation with Fun Libs allows comparison of our approach with that proposed by Hossain et al.,~\shortcite{Hossain2017}.\footnote{We limit ourselves to the 50 stories created by Hossain et al.,~\shortcite{Hossain2017}; we believe that ours is the first endeavor in uncovering the demographic-specific idiosyncrasies in humor preferences, and provides motivation for future efforts to collect even larger demographic-specific datasets for humor studies.}
We discard 4 of the 50 Fun Libs, as their themes cater to a US audience,\footnote{Their titles include {\it Kim Kardashian}, {\it Baseball}, {\it Boston Tea Party}, {\it The Statue of Liberty}.} and replace them with 4 new stories we created following the heuristics devised by Hossain et al.,~\shortcite{Hossain2017}.

The data annotation is undertaken by two parties: the {\bf players} who fill-in the blanks to create funny stories, and the {\bf judges} who assess the filled-in stories in terms of their funniness.
We assume the three-stage annotation framework devised by Hossain et al.,~\shortcite{Hossain2017}: {\it judge selection}, {\it player selection} and {\it story annotation}, with a few revisions to account for location-specific annotations.
\paragraph{Judge selection.}
This is done via a linguistic and a demographic survey.
Turker judges from each country are given seven pre-filled stories, three of which are taken directly from Wikipedia, with some words underlined as if they were filled-in words, while the remaining four stories were filled-in by English speakers from the corresponding country, who were instructed to create funny stories.
We instruct the judges to select for each story a grade from \{0: not funny, 1: slightly funny, 2: moderately funny, 3: funny\}.
To filter spam responses, verification questions are presented that can be answered only after reading the stories.
The task ends with a demographic survey prompting the judges to provide their age group, nationality, gender, education, occupation, and income level.

Those turkers are selected who (i) assign 0 to the Wikipedia stories, a grade from \{1, 2, 3\} to at least three of the remaining four stories, (ii) answer the story-based questions correctly, and (iii) spend at least 4 minutes to complete the task.
50 US and 43 IN judges are selected from 60 and 100 candidates respectively, ensuring that they are unbiased in judging the funniness of a filled-in story.
\paragraph{Player selection.}
Players are expected to be good at writing funny stories.
For this, we obtain four stories from Wikipedia, mask some words, and provide hint types next to them.
To avoid excess workload to the turkers (leading to possible filling bias), two tasks are created for each country on AMT, each with two stories and a demographic survey, with instructions to fill the stories to make them funny, and answer the demographic survey.
Further instructions are provided: the filled-in words must (i) 
occur in English dictionary, 
(ii) have exactly one word written in Latin alphabet, (iii) agree with the provided hint types, and (iv) not be slang words or sexual references, as these lead to shallow humor. 

In each country, the filled-in stories are graded for humor on the 0-3 scale by 5 qualified judges from that country, to lessen the effect of variations in humor preferences, and be representative of an audience rather than an individual on AMT.
Players are selected if their {\bf mean funniness grade ({\sc mfg})}, mean of the 5 judgements, $\geq$ 1 for at least one story.
30 IN and 26 US players are selected from 80
and 60 turkers respectively.
\paragraph{Annotating Fun Libs.}
For each story, we ensure that a qualified turker can participate as either a player or a judge but not both.
The stories are filled by 3 players from each country following the player selection instructions, and are judged by 5 judges from the same country.
The judges rate the overall funniness, coherence, deviation from the story title (on the 0-3 scale), whether incongruity\footnote{Incongruity is present in a joke if there is a surprise that defies the expectation of the reader \cite{Weems2014}, therefore causing the content to be funny.} was applied by the player (\{Yes, No\}), the humor contribution of each filled-in word (\{`funny', `not funny'\}), and a verification question.
In addition to same country judgements, we also obtain judgements from the opposite country for the filled-in test stories, to allow for cross-country analyses.
We compare coherence to incongruity and deviation, to understand their effect on humor.
The Krippendorf's alpha \cite{Krippendorff1970} values for IN and US are $0.214$ and $0.173$ respectively, which indicate positive agreements among AMT judges, and are comparable to those obtained by Hossain et al.~\shortcite{Hossain2019}, who crowd-sourced a dataset of humorous edited news headlines on the same funniness scale.
The total AMT cost is about \$1,200.

\section{Automatic Humor Generation Framework}
\label{sec:framework}
We first describe the location-specific training components, namely the location-biased language model 
and the location-specific humor classifier. 
Then, we focus on the story generation pipeline that follows three stages: (1) A {\bf candidate selection} stage where possible word replacements for the blanks are generated using a location-biased language model; (2) A {\bf candidate ranking} stage where the selected candidates are ranked by their humor contribution to the sentences they occur in, using a location-specific humor classifier; (3) A {\bf story completion} stage where funny stories are created by selecting the top ranked funny transformations for each sentence, and concatenating them to obtain complete stories.


\subsection{Training components}
\subsubsection{Location-Biased Language Model}
In order to be able to generate a Mad Lib-like story, the first step is to automatically fill in a blank with a word that fits the context. Such words can be predicted by a language model, such as the BERT masked language model (MLM), a state-of-the-art deep learning framework \cite{Devlin2019} based on a multi-layer bidirectional Transformer \cite{Vaswani2017}, trained on the English Wikipedia and Book Corpus \cite{Zhu2015} datasets, for masked word and next sentence prediction tasks. Since its predictive ability is generic and it does not take the desired demographics into consideration (such as location, in our case), we train it further on location-rich data, thus enabling the model to make word predictions in context biased toward a particular country. To achieve this, we use a large dataset of blog posts \cite{Garimella2017} authored by users from IN and US (35K blogs, 17M tokens for IN, and 33K, 12M tokens for US). We use BERT$_{\mathrm{base}}$ model with default parameters.
This allows the language model to incorporate location-based word preferences in its prediction and to provide different replacements for a masked word occurring in a sentence written by an Indian English speaker versus an American English speaker. 



\subsubsection{Location-Specific Humor Classification}
Furthermore, automatically generated word replacements in context need to be assessed for humor given an audience with a particular demographic. To enable us to gauge the funniness level of a word replacement, we train location-specific humor classifiers based on the BERT framework by leveraging the AMT annotations and the country of the turkers.
Of the 50 stories, 40 are used for training,\footnote{The 4 newly created stories replace 4 training set stories.} and the remaining for evaluation. The stories' sentence splits are retained from 
\cite{Hossain2017}. The training and validation datasets consist of sentence pairs and their associated humor labels which are derived as follows. 
First, labels are assigned to each filling using a {\bf majority vote} over the funny judgements in the gold standard. A sentence is considered {\bf funny} if at least 50\% of the filled-in words are funny. Sentences that do not contain blanks are not used for training.

\begin{table}[t]
\centering
\scalebox{0.85}{
\begin{tabular}{l|c|c|c|c}
\toprule
\textsc{Type} & \multicolumn{2}{c|}{\textsc{Funny}} & \multicolumn{2}{c}{\textsc{Not Funny}} \\
\midrule
& \textsc{IN} & \textsc{US} & \textsc{IN} & \textsc{US} \\
\midrule
\textsc{Train} & 566 & 574 & 130 & 122\\
\textsc{Validation} & 173 & 193 & 49 & 29 \\
\textsc{Test} & 137 & 210 & 94 & 21\\
\bottomrule
\end{tabular}
}
\caption{Statistics on the humor classification datasets.}
\label{tab:humorDataStats}
\end{table}

Table \ref{tab:humorDataStats} shows the sizes the train, validation and test sets.
Since we instructed the players to create funny stories, the datasets are, as expected, skewed toward the funny class. We opt to augment our data with additional non-funny sentence-pairs from Wikipedia. Since candidate sentences are location-agnostic, and therefore would have the same form for both India and US, we introduce location biased completions.  We use our location-specific MLM to replace masked words in Wikipedia sentences\footnote{We focus on one of the four POS tags, as other hint types are not trivial to identify.} with the highest probability word given the location model, resulting in non-funny sentence pairs that are different by location.

We use the above augmented dataset to fine-tune {\textsc{FunnyBERT}}, a BERT-framework based sentence humor classifier that, accounting for the desired country of the audience, is able to identify whether content is humorous or not. 
Specifically, we fine-tune BERT for sentence-pair classification by adding a classification layer, with input pair $<$masked sentence, filled-in sentence$>$, and output is the prediction $c \in$ \{funny, not funny\} if the input corresponds to a funny transformation.
The final hidden vector $h$  corresponding to [\texttt{CLS}] (classification token) represents the sequence.
The weights $W$ are learnt during fine-tuning; $p(c|h)=\mathrm{softmax}(Wh)$.
We use the following parameters: batch size 32, \texttt{gelu} activation, sequence length 512, vocabulary 30,522, Adam optimizer, learning rate 1e-5, (selected over 5e-5, 5e-6, 1e-6), and 10 epochs (over 1-100 epochs). 
The limited sizes of annotated datasets make BERT a suitable framework to build on.
FunnyBERT ranks the candidate filled-in sentences based on their humor (softmax probability).

\begin{table}[t]
\centering
\scalebox{0.9}{
\begin{tabular}{p{9cm}|l|l}
\toprule
\textsc{Metric} & \textsc{IN} & \textsc{US}\\
\midrule
{\sc Precision (Funny)} & 81.48 & 91.16$^*$ \\
{\sc Recall (Funny)} & 89.02$^*$ & 85.49 \\
{\sc F1 score (Funny)} & 85.08 & 88.24$^*$ \\
{\sc Accuracy} & {\bf 84.39} & {\bf 88.60}$^*$ \\
{\sc Accuracy} (Wikipedia sentences without modification) & 80.06 & {\bf 89.12}$^*$ \\
\bottomrule
\end{tabular}
}
\caption{Classification validation accuracies of {\sc FunnyBERT$_{\mathrm{X}}$}, $\mathrm{X} \in$ \{IN, US\} ($p < 0.05$).}
\label{tab:humorClassificationValidation}
\vspace{-0.1in}
\end{table}

Table \ref{tab:humorClassificationValidation} shows the location-specific validation accuracies of FunnyBERT.
The majority vote baseline accuracy is $50\%$, as the datasets are class-balanced.
Metric-wise statistically significant values are marked with $^*$. 
The funny-class precision and accuracy are lower, and recall is higher for IN than those for US.
This may be due to slightly lower quality funny sentence completions from IN players, resulting in FunnyBERT$_{\mathrm{IN}}$ predicting false positives in most cases. 
We suspect that the familiarity of US turkers with Mad Libs was a factor that led to better quality stories for US. 
Augmenting Wikipedia sentences without location-specific replacements results in a significantly lower accuracy ($80.06\%$) for IN, suggesting that the improved US scores are caused by the biased nature of the pre-training datasets used by BERT toward US English;  IN-specific replacements, however minor, lead to improved performances in this locale.

\subsection{{\sc YodaLib} for Story Generation}
Here we introduce YodaLib, our pipeline for humorous story generation. For a given MadLib-like story, we start out with word candidate selection and ranking, and finalize with story completion.

\subsubsection{Candidate Selection}

In order to generate candidate words for each story blank, 
the latter are replaced with the BERT [MASK] token 
and then input to the location-specific MLM.
For each mask, probability scores are obtained for all the words in the vocabulary, and we retain the highest ranking $k=10,000$ words.
These are further filtered to obtain a cleaner candidate list, adhering to hint types and other restrictions imposed in the player selection phase; we further ensure that the number and tense match the hint type for nouns and verbs, respectively. 

In sentences with multiple masks, we perform left-to-right selection, at each step pruning our decision space to the top $n$ candidates to avoid assessing an exponential number of combinations.   

We use {\sc FunnyBERT} to rank the filled-in sentences (explained in Section \ref{sec:candidateRanking}),
and the top $n=100$ candidates are chosen to fill each mask.
We impose left-to-right candidate selection, as this is how stories are read by turkers.
It enhances the overall coherence of the resulting funny sentence, as the previously selected candidates are considered in selecting next ones.
For a given context, we expect the candidates with high MLM scores to be good fits and hence less humorous ({\it cats \underline{drink} milk}), while those with low scores are more likely to be incongruous, and may generate humor in the given context ({\it cats \underline{prepare} milk}).
\subsubsection{Candidate Ranking}
\label{sec:candidateRanking}
Examining the positions or scores from MLM is not sufficient to predict if candidates are funny substitutes.
The second stage involves ranking the candidates for each mask based on their humor contributions to the containing sentences. We leverage the FunnyBERT component by feeding the completed senteces to the model and using the softmax humor probability as a ranking value. The top $n=100$ sentences are used for candidate selection of the next mask, until all the masks are filled.

\subsubsection{Story Completion}
The final stage of the framework involves forming complete stories from the top funny transformations for each of the component masked sentences.
Similar to candidate selection, we consider left-to-right story completion. For a sentence to be appended, it must be classified both as funny and as a potential next sentence given the previously selected context. 
This allows the resulting stories to be both funny and coherent. 
For (1), we consider the top-ranked funny transformations for each sentence.
For (2), we rank the funny transformations based on their semantic similarity to the sentences previously selected in the story. 
The top $N=100$ funny {\it and similar} transformations are selected for each subsequent sentence, resulting in different variations of the completed story.
Only the top $N$ as yet completed stories are advanced after processing each sentence (from $N^2$ sequences), based on the above two scores. 

We estimate the similarity between any two transformations as the cosine similarity between their sentence embeddings.
We use the average of the word embeddings from location-specific BERT to estimate the sentence embedding. 
Each word has 12 vectors (from the 12 BERT layers), each of length 768.
Different layers of BERT encode different kinds of information, so the appropriate aggregation strategy depends on the task.
We consider two variations to obtain the final word vectors: (i) sum of the embeddings from the last 4 hidden layers, and (i) embeddings from only the second to last hidden layer \cite{Devlin2019}.
We settle on the second strategy, 
as it results in better quality stories for the validation set.\footnote{The quality of stories is determined by human subjects, as this task is subjective.}
We then sort the stories in decreasing order of their {\bf story funniness score} (mean of funniness scores of the constituting sentences) 
, as well as according to their {\bf average word coherence} (mean of the pair-wise similarities of filled-in word embeddings).

\section{Evaluation}
\label{sec:evaluation}
 We evaluate three methods to write funny stories:
\begin{enumerate}\itemsep-0.2em 
    \item \textbf{\textsc{FreeText (FT)}}: Following directions, AMT players complete the funny stories.
    \item \textbf{\textsc{MLM}}: Word-replacement for each blank is selected based on the original BERT MLM model (without location-specific training).
    \item \textbf{\textsc{YodaLib}}: The stories are generated using the proposed framework (see Section \ref{sec:framework}).
\end{enumerate}
\noindent For each of these, we obtain 5 judgements on AMT. For each test Fun Lib, grading is done by judges from both locations on the 3 filled-in stories from {\sc FT}, and the top 10 stories generated from {\sc MLM} and {\sc YodaLib}.
We grade more stories generated by {\sc YodaLib} as: (1) unlike the manually crafted {\sc FT} stories , these are generated automatically, and hence require minimal effort (once the YodaLib framework is in place), and (2) both story funniness and average word coherence scores for the top 10 stories differ only in the fourth decimal place.
Hence we treat them as the best diverse humorous variations generated for the given Fun Libs ({\sc Best10}).
The stories from \textsc{FT} and \textsc{YodaLib} approaches are created in location-specific settings, while \textsc{MLM} does not account for the desired location slant.
Stories from {\sc MLM} are ranked based on their filled-in word probabilities of fitting the corresponding contexts.
Hence, we expect the top-ranked {\sc MLM} stories to be more congruous and less humorous than the low-ranked stories.

\begin{table}[t]
\centering
\scalebox{0.85}{
\begin{tabular}{p{1.6cm}|l|l|l|l}
\toprule
\textsc{Method} & \multicolumn{2}{c|}{\textsc{IN judges}} & \multicolumn{2}{c}{\textsc{US judges}} \\
\midrule
& \textsc{Top3} & \textsc{Top10} & \textsc{Top3} & \textsc{Top10} \\
\midrule
{\sc FT}$_{\mathrm{IN}}$ & 1.17$^\ddagger$ & - & \underline{{1.39}}$^{\ddagger\mathparagraph}$ & - \\
{\sc FT}$_{\mathrm{US}}$ & \underline{1.57}$^{*\ddagger}$ & - & 1.41$^\ddagger$ & - \\
{\sc MLM} & 0.70 & 0.91 & 0.68 & 0.84 \\
{\sc Yoda}$_{\mathrm{IN}}$ & \underline{{\textbf{1.94}}}$^{*\dagger\ddagger\mathparagraph}$ & \underline{{1.60}}$^{*\ddagger\mathparagraph}$ & 1.56$^{*\ddagger}$ & 1.32$^{\ddagger}$ \\
{\sc Yoda}$_{\mathrm{US}}$ & \underline{{\textbf{2.03}}}$^{*\dagger\ddagger\mathparagraph}$ & \underline{{\textbf{1.70}}}$^{*\dagger\ddagger\mathsection\mathparagraph}$ & {\bf 1.77}$^{*\dagger\ddagger\mathsection}$ & 1.48$^{*\ddagger\mathsection}$  \\
\bottomrule
\end{tabular}
}
\caption{Average {\sc mfg} for generated stories. Column-wise significantly higher scores than (i) {\sc FT}$_{\mathrm{IN}}$ are marked with $^*$, (ii) {\sc FT}$_{\mathrm{US}}$ with $^\dagger$, (iii) {\sc MLM} with $^\ddagger$, and (iv) {\sc Yoda}$_{\mathrm{IN}}$ with $^\mathsection$. 
Row-wise higher values for each country are marked with $^\mathparagraph$ ($p < 0.05$). The highest values in columns/ rows are bold/ underlined. 
}
\label{tab:meanFunninessGrades}
\end{table}
Table \ref{tab:meanFunninessGrades} shows the averages of the mean funniness grades {\sc mfg} (0-3 scale) of the generated stories, in two settings (for {\sc MLM} and {\sc Yoda}).
(1) {\sc Top3}: For {\sc MLM}, we consider top three stories for each Fun Lib. For {\sc YodaLib}, we consider three stories from {\sc Best10} which have the highest {\sc mfg} assigned by judges (as all of them have very similar scores assigned by our framework).
(2) {\sc Top10}: We consider the top 10 (from filled-in word scores) and {\sc Best10} stories for each Fun Lib for {\sc MLM} and {\sc YodaLib} respectively.

Humor generated by both {\sc FT} and {\sc YodaLib} approaches is preferred to that by {\sc MLM} by both IN and US judges.
This is expected, as {\sc MLM} fills the stories with more plausible words, and hence does not introduce any surprise aspect that is essential for humor.
The average {\sc mfg} for {\sc MLM} increases in the {\sc Top10} setting, confirming that stories become more humorous as more incongruous words are used to fill them.
In the {\sc FT} stories, both IN and US judges prefer US-written humor ($1.57$, $1.41$) to IN-written humor ($1.17$, $1.39$). 
On the other hand, IN-written humor is liked more by US judges ($1.39$), and US-written humor by IN judges ($1.57$). 
Hence, an average US turker writes better stories than an average IN turker for Mad Libs, and judges in general enjoy humor written by turkers from a different country more than of their countrymen.
This is in contrast to the general expectation that people prefer humor originated from their own group due to a better understanding of the various location-specific subtleties, suggesting that seeing things in a new light possibly contributes to the surprise and creativity factors essential for humor generation.

In the {\sc YodaLib} approach, US-written humor is preferred to IN-written humor by both IN ($2.03$, $1.70$) and US ($1.77$, $1.48$) judges, in both settings.
This may be due to the better performance of FunnyBERT$_{\mathrm{US}}$ in ranking funny transformations to create stories.
As seen in the {\sc FT} approach, US-written stories by turkers are of better quality humor than IN-written ones on average, and this may have led to better quality training datsets for FunnyBERT$_{\mathrm{US}}$.
Our {\sc YodaLib} approach outperforms {\sc FT} for both IN and US judges (more so in the {\sc Top3} setting for US judges, and both settings for IN judges), indicating that our approach generates better quality humor than AMT players on an average.
Our approach also outperforms Libitum \cite{Hossain2017}
in the {\sc Top3} setting.
 It has an average {\sc mfg} of $1.51$, which is much lower than the $1.77$  given by US judges to US-generated humor.
\footnote{Training and evaluation settings for Libitum are closer to US humor judged by US judges setting ({\sc Top3}).}
 
 Both the IN- and US-generated stories in the {\sc YodaLib} approach are liked more by IN judges than by US judges (underlined in Table \ref{tab:meanFunninessGrades}).
This suggests that an average IN turker has a more lenient outlook towards humor, whether it is of IN or US origin, and this may have also lead to lower quality IN-written stories, whereas an average US turker 
a more stricter perspective towards it, possibly due to the familiarity of the game in US, resulting in the more enjoyable humor from US players.

\section{Discussion}
\label{sec:discussion}
\begin{table}[!htbp]
\centering
\scalebox{0.85}{
\begin{tabular}{p{1.4cm}|c|r|r|r|r|r}
\toprule
\textsc{Method} & \multicolumn{3}{c|}{{\sc Indian judges}} & \multicolumn{3}{c}{{\sc US judges}} \\
\midrule
 & {\sc Coh} & {\sc Inc} & {\sc Dev} & {\sc Coh} & {\sc Inc} & {\sc Dev} \\
 \midrule
{\sc FT}$_{\mathrm{IN}}$ & {\bf 0.40} & 0.25 & {\bf 0.59} & 0.01 & {\bf 0.68} & 0.19\\
{\sc FT}$_{\mathrm{US}}$ & 0.23 & 0.24 & {\bf 0.62} & {\bf 0.77} & {\bf 0.83} & {\bf -0.59}\\
{\sc MLM} & {\bf -0.57} & {\bf 0.74} & {\bf 0.66} & {\bf -0.64} & {\bf 0.36} & {\bf 0.72}\\
{\sc Yoda}$_{\mathrm{IN}}$ & {\bf 0.22} & {\bf 0.41} & {\bf 0.52} & 0.15 & 0.12 & {\bf 0.22}\\
{\sc Yoda}$_{\mathrm{US}}$ & 0.07 & {\bf 0.27} & {\bf 0.41} & 0.12 & {\bf 0.27} & 0.05\\
\bottomrule
\end{tabular}
}
\caption{Correlations of coherence (Coh), incongruity (Inc) and deviation (Dev) values with {\sc mfg}. Statistically significant values are in {\bf bold} ($p < 0.05$).}
\label{tab:humorCorrelations}
\end{table}
Table \ref{tab:humorCorrelations} shows the correlations of coherence, incongruity and deviation with the corresponding {\sc mfg} for the test stories generated in each approach.
The {\sc Top10} stories are used for {\sc MLM} and {\sc YodaLib} approaches.
In {\sc FT}, coherence plays an important role in creating humor for same-location audience on AMT (IN: $0.40$, US: $0.77$), indicating that humans have a striking ability to apply coherence to create humor, and it is particularly appreciated by audience from the same country, possibly due to a consistent use of location-specific information.
This can be seen in the US judgements on IN-written humor, where funniness has no correlation with coherence, and high correlation with incongruity, indicating that US judges find those IN-written stories funny which contain seemingly unexpected words.

In the {\sc MLM} stories, coherence is negatively correlated with funniness, indicating the difficulty in generating humor via coherent words by a language model.
In the {\sc YodaLib} approach, coherence plays very little role; most of the funniness is achieved via incongruity and topic deviations (the only exception being US judgements on US-written stories for deviation), though IN judges find IN-generated humor to be somewhat coherent ($0.22$).
Certain skilled turkers are able to write meaningful and coherent stories, something that the {\sc YodaLib} approach finds very difficult to achieve.

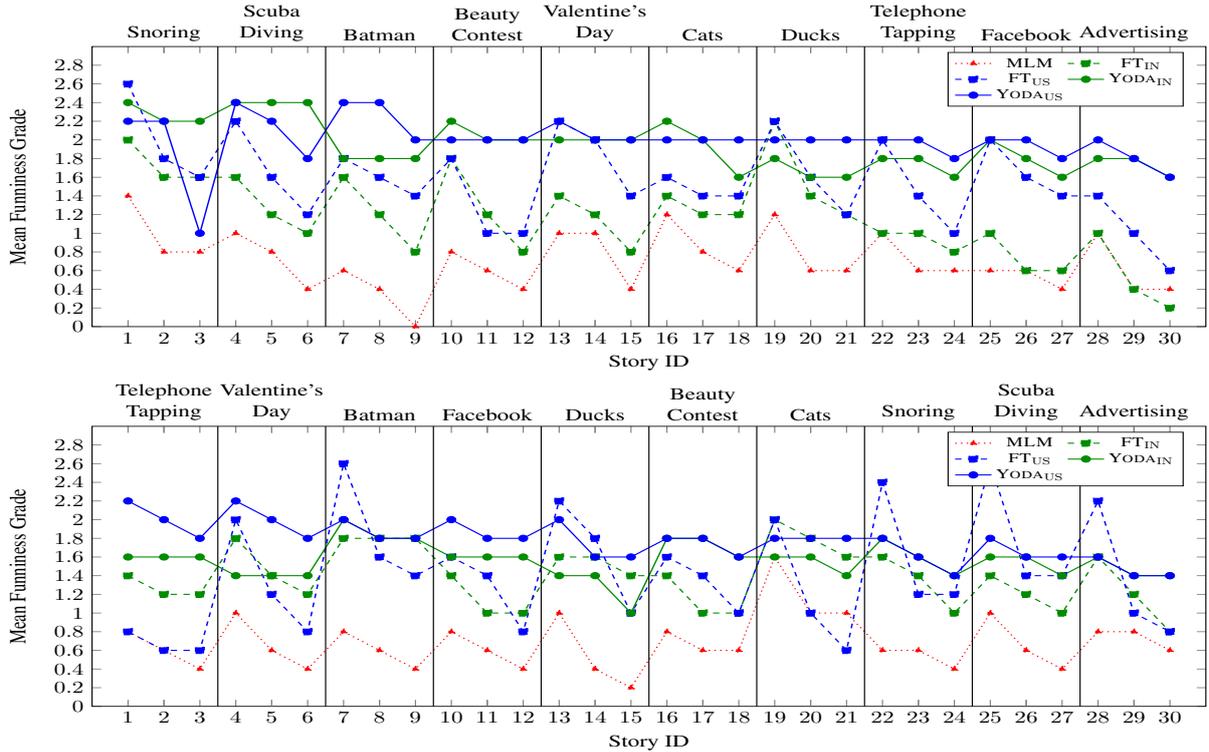
\begin{figure*}[t]
\resizebox{\linewidth}{5cm} {
\begin{tikzpicture}
    \begin{axis}[
        width=20cm,height=8cm,
        xlabel=Story ID,
        ylabel=Mean Funniness Grade,
        legend columns=2,
        legend style={nodes={scale=0.8, transform shape}},
        xmin=0,ymin=0,
        xmax=31,ymax=3.0,
        ytick={0,0.2,0.4,0.6,0.8,1.0,1.2,1.4,1.6,1.8,2.0,2.2,2.4,2.6,2.8},
        xtick={1,2,3,4,5,6,7,8,9,10,11,12,13,14,15,16,17,18,19,20,21,22,23,24,25,26,27,28,29,30},
        extra x ticks={2,5,8,11,14,17,20,23,26,29},
extra x tick style={grid=none,ticks=major,ticklabel pos=right,align=center},
extra x tick labels={Snoring,Scuba\\Diving,Batman,Beauty\\Contest,Valentine's\\Day,Cats,Ducks,Telephone\\Tapping,Facebook,Advertising}
        ]
    \addplot[dotted,mark=triangle*,red,line width=0.7] plot coordinates {
        (1, 1.4)
        (2, 0.8)
        (3, 0.8)
        (4, 1.0)
        (5, 0.8)
        (6, 0.4)
        (7, 0.6)
        (8, 0.4)
        (9, 0.0)
        (10, 0.8)
        (11, 0.6)
        (12, 0.4)
        (13, 1.0)
        (14, 1.0)
        (15, 0.4)
        (16, 1.2)
        (17, 0.8)
        (18, 0.6)
        (19, 1.2)
        (20, 0.6)
        (21, 0.6)
        (22, 1.0)
        (23, 0.6)
        (24, 0.6)
        (25, 0.6)
        (26, 0.6)
        (27, 0.4)
        (28, 1.0)
        (29, 0.4)
        (30, 0.4)
    };
    \addlegendentry{\textsc{MLM}}
    \addplot[dashed,mark=square*,mynicegreen,line width=0.7] plot coordinates {
         (1, 2.0)
        (2, 1.6)
        (3, 1.6)
        (4, 1.6)
        (5, 1.2)
        (6, 1.0)
        (7, 1.6)
        (8, 1.2)
        (9, 0.8)
        (10, 1.8)
        (11, 1.2)
        (12, 0.8)
        (13, 1.4)
        (14, 1.2)
        (15, 0.8)
        (16, 1.4)
        (17, 1.2)
        (18, 1.2)
        (19, 2.2)
        (20, 1.4)
        (21, 1.2)
        (22, 1.0)
        (23, 1.0)
        (24, 0.8)
        (25, 1.0)
        (26, 0.6)
        (27, 0.6)
        (28, 1.0)
        (29, 0.4)
        (30, 0.2)
    };
    \addlegendentry{\textsc{FT}$_{\mathrm{IN}}$}
    \addplot[dashed,mark=square*,blue,line width=0.7] plot coordinates {
         (1, 2.6)
        (2, 1.8)
        (3, 1.6)
        (4, 2.2)
        (5, 1.6)
        (6, 1.2)
        (7, 1.8)
        (8, 1.6)
        (9, 1.4)
        (10, 1.8)
        (11, 1)
        (12, 1)
        (13, 2.2)
        (14, 2.0)
        (15, 1.4)
        (16, 1.6)
        (17, 1.4)
        (18, 1.4)
        (19, 2.2)
        (20, 1.6)
        (21, 1.2)
        (22, 2.0)
        (23, 1.4)
        (24, 1.0)
        (25, 2.0)
        (26, 1.6)
        (27, 1.4)
        (28, 1.4)
        (29, 1.0)
        (30, 0.6)
    };
    \addlegendentry{\textsc{FT}$_{\mathrm{US}}$}
    \addplot[solid,mark=*,mynicegreen,line width=0.7] plot coordinates {
        (1, 2.4)
        (2, 2.2)
        (3, 2.2)
        (4, 2.4)
        (5, 2.4)
        (6, 2.4)
        (7, 1.8)
        (8, 1.8)
        (9, 1.8)
        (10, 2.2)
        (11, 2.0)
        (12, 2.0)
        (13, 2.0)
        (14, 2.0)
        (15, 2.0)
        (16, 2.2)
        (17, 2.0)
        (18, 1.6)
        (19, 1.8)
        (20, 1.6)
        (21, 1.6)
        (22, 1.8)
        (23, 1.8)
        (24, 1.6)
        (25, 2.0)
        (26, 1.8)
        (27, 1.6)
        (28, 1.8)
        (29, 1.8)
        (30, 1.6)
    };
    \addlegendentry{\textsc{Yoda}$_{\mathrm{IN}}$}
     \addplot[solid,mark=*,blue,line width=0.7] plot coordinates {
        (1, 2.2)
        (2, 2.2)
        (3, 1.0)
        (4, 2.4)
        (5, 2.2)
        (6, 1.8)
        (7, 2.4)
        (8, 2.4)
        (9, 2.0)
        (10, 2.0)
        (11, 2.0)
        (12, 2.0)
        (13, 2.2)
        (14, 2.0)
        (15, 2.0)
        (16, 2.0)
        (17, 2.0)
        (18, 2.0)
        (19, 2.0)
        (20, 2.0)
        (21, 2.0)
        (22, 2.0)
        (23, 2.0)
        (24, 1.8)
        (25, 2.0)
        (26, 2.0)
        (27, 1.8)
        (28, 2.0)
        (29, 1.8)
        (30, 1.6)
    };
    \addlegendentry{\textsc{Yoda}$_{\mathrm{US}}$}
    \addplot [solid,mark=.,black,line width=0.5]
    coordinates {(3.5, 0) (3.5, 3)};
    \addplot [solid,mark=.,black,line width=0.5]
    coordinates {(6.5, 0) (6.5, 3)};
    \addplot [solid,mark=.,black,line width=0.5]
    coordinates {(9.5, 0) (9.5, 3)};
    \addplot [solid,mark=.,black,line width=0.5]
    coordinates {(12.5, 0) (12.5, 3)};
    \addplot [solid,mark=.,black,line width=0.5]
    coordinates {(15.5, 0) (15.5, 3)};
    \addplot [solid,mark=.,black,line width=0.5]
    coordinates {(18.5, 0) (18.5, 3)};
    \addplot [solid,mark=.,black,line width=0.5]
    coordinates {(21.5, 0) (21.5, 3)};
    \addplot [solid,mark=.,black,line width=0.5]
    coordinates {(24.5, 0) (24.5, 3)};
    \addplot [solid,mark=.,black,line width=0.5]
    coordinates {(27.5, 0) (27.5, 3)};
    \end{axis}
    \end{tikzpicture}
    }
    \label{fig:meanFunninessGradesIndia}
\resizebox{\linewidth}{5cm} {
\begin{tikzpicture}
    \begin{axis}[
        width=20cm,height=8cm,
        xlabel=Story ID,
        ylabel=Mean Funniness Grade,
        legend columns=2,
        legend style={nodes={scale=0.8, transform shape}},
        xmin=0,ymin=0,
        xmax=31,ymax=3.0,
        ytick={0.0,0.2,0.4,0.6,0.8,1.0,1.2,1.4,1.6,1.8,2.0,2.2,2.4,2.6,2.8},
        xtick={1,2,3,4,5,6,7,8,9,10,11,12,13,14,15,16,17,18,19,20,21,22,23,24,25,26,27,28,29,30},
        extra x ticks={2,5,8,11,14,17,20,23,26,29},
extra x tick style={grid=none,ticks=major,ticklabel pos=right,align=center},
extra x tick labels={Telephone\\Tapping, Valentine's\\Day,Batman,Facebook,Ducks,Beauty\\Contest,Cats,Snoring,Scuba\\Diving,Advertising}
        ]
    \addplot[dotted,mark=triangle*,red,line width=0.7] plot coordinates {
        (1, 0.8)
        (2, 0.6)
        (3, 0.4)
        (4, 1.0)
        (5, 0.6)
        (6, 0.4)
        (7, 0.8)
        (8, 0.6)
        (9, 0.4)
        (10, 0.8)
        (11, 0.6)
        (12, 0.4)
        (13, 1.0)
        (14, 0.4)
        (15, 0.2)
        (16, 0.8)
        (17, 0.6)
        (18, 0.6)
        (19, 1.6)
        (20, 1.0)
        (21, 1.0)
        (22, 0.6)
        (23, 0.6)
        (24, 0.4)
        (25, 1.0)
        (26, 0.6)
        (27, 0.4)
        (28, 0.8)
        (29, 0.8)
        (30, 0.6)
    };
    \addlegendentry{\textsc{MLM}}
    \addplot[dashed,mark=square*,mynicegreen,line width=0.7] plot coordinates {
         (1, 1.4)
        (2, 1.2)
        (3, 1.2)
        (4, 1.8)
        (5, 1.4)
        (6, 1.2)
        (7, 1.8)
        (8, 1.8)
        (9, 1.8)
        (10, 1.4)
        (11, 1.0)
        (12, 1.0)
        (13, 1.6)
        (14, 1.6)
        (15, 1.4)
        (16, 1.4)
        (17, 1.0)
        (18, 1.0)
        (19, 2.0)
        (20, 1.8)
        (21, 1.6)
        (22, 1.6)
        (23, 1.4)
        (24, 1.0)
        (25, 1.4)
        (26, 1.2)
        (27, 1.0)
        (28, 1.6)
        (29, 1.2)
        (30, 0.8)
    };
    \addlegendentry{\textsc{FT}$_{\mathrm{IN}}$}
    \addplot[dashed,mark=square*,blue,line width=0.7] plot coordinates {
         (1, 0.8)
        (2, 0.6)
        (3, 0.6)
        (4, 2.0)
        (5, 1.2)
        (6, 0.8)
        (7, 2.6)
        (8, 1.6)
        (9, 1.4)
        (10, 1.6)
        (11, 1.4)
        (12, 0.8)
        (13, 2.2)
        (14, 1.8)
        (15, 1.0)
        (16, 1.6)
        (17, 1.4)
        (18, 1.0)
        (19, 2.0)
        (20, 1.0)
        (21, 0.6)
        (22, 2.4)
        (23, 1.2)
        (24, 1.2)
        (25, 2.6)
        (26, 1.4)
        (27, 1.4)
        (28, 2.2)
        (29, 1.0)
        (30, 0.8)
    };
    \addlegendentry{\textsc{FT}$_{\mathrm{US}}$}
    \addplot[solid,mark=*,mynicegreen,line width=0.7] plot coordinates {
        (1, 1.6)
        (2, 1.6)
        (3, 1.6)
        (4, 1.4)
        (5, 1.4)
        (6, 1.4)
        (7, 2.0)
        (8, 1.8)
        (9, 1.8)
        (10, 1.6)
        (11, 1.6)
        (12, 1.6)
        (13, 1.4)
        (14, 1.4)
        (15, 1.0)
        (16, 1.8)
        (17, 1.8)
        (18, 1.6)
        (19, 1.6)
        (20, 1.6)
        (21, 1.4)
        (22, 1.8)
        (23, 1.6)
        (24, 1.4)
        (25, 1.6)
        (26, 1.6)
        (27, 1.4)
        (28, 1.6)
        (29, 1.4)
        (30, 1.4)
    };
    \addlegendentry{\textsc{Yoda}$_{\mathrm{IN}}$}
     \addplot[solid,mark=*,blue,line width=0.7] plot coordinates {
        (1, 2.2)
        (2, 2.0)
        (3, 1.8)
        (4, 2.2)
        (5, 2.0)
        (6, 1.8)
        (7, 2.0)
        (8, 1.8)
        (9, 1.8)
        (10, 2.0)
        (11, 1.8)
        (12, 1.8)
        (13, 2.0)
        (14, 1.6)
        (15, 1.6)
        (16, 1.8)
        (17, 1.8)
        (18, 1.6)
        (19, 1.8)
        (20, 1.8)
        (21, 1.8)
        (22, 1.8)
        (23, 1.6)
        (24, 1.4)
        (25, 1.8)
        (26, 1.6)
        (27, 1.6)
        (28, 1.6)
        (29, 1.4)
        (30, 1.4)
    };
    \addlegendentry{\textsc{Yoda}$_{\mathrm{US}}$}
    \addplot [solid,mark=.,black,line width=0.5]
    coordinates {(3.5, 0) (3.5, 3)};
    \addplot [solid,mark=.,black,line width=0.5]
    coordinates {(6.5, 0) (6.5, 3)};
    \addplot [solid,mark=.,black,line width=0.5]
    coordinates {(9.5, 0) (9.5, 3)};
    \addplot [solid,mark=.,black,line width=0.5]
    coordinates {(12.5, 0) (12.5, 3)};
    \addplot [solid,mark=.,black,line width=0.5]
    coordinates {(15.5, 0) (15.5, 3)};
    \addplot [solid,mark=.,black,line width=0.5]
    coordinates {(18.5, 0) (18.5, 3)};
    \addplot [solid,mark=.,black,line width=0.5]
    coordinates {(21.5, 0) (21.5, 3)};
    \addplot [solid,mark=.,black,line width=0.5]
    coordinates {(24.5, 0) (24.5, 3)};
    \addplot [solid,mark=.,black,line width=0.5]
    coordinates {(27.5, 0) (27.5, 3)};
    \end{axis}
    \end{tikzpicture}
    }
    \caption{{\sc mfg} for the 30 test stories (top: IN, bottom: US) using the three approaches graded by US judges.}
    \label{fig:meanFunninessGradesUS}
\end{figure*}
\subsection{Qualitative Analysis}
Figure \ref{fig:meanFunninessGradesUS} shows {\sc mfg} for stories via the three approaches by IN and US judges respectively.
We sort the story titles in decreasing order of {\sc mfg}.
{\sc FT} consistently outperforms {\sc MLM}.
{\sc YodaLib} outperforms {\sc FT} for 28 and 23 stories for IN and US judges respectively.
The average gains are $0.79$ and $0.46$ for IN judges, and $0.17$ and $0.37$ for US judges, for IN- and US-generated stories respectively.
Hence, an average IN judge, though also prefers coherence, finds stories with low coherence and high incongruity and deviation also funny; as these are further incorporated in {\sc YodaLib} stories, the gains in IN judgements are higher.
Contrarily, US judges prefer coherence along with incongruity:
the highest {\sc mfg} by IN judges are for {\sc YodaLib} stories (e.g., {\it Snoring}, {\it Scuba Diving}, {\it Batman}); those by US judges are for {\sc FT} stories (e.g., {\it Batman}, {\it Snoring}, {\it Scuba Diving}).
It is interesting that the original Mad Libs game introduced humor primarily via incongruous words to fill the stories, while coherence plays an important role in this study, possibly due to the richness of the task in considering the context to fill-in the blanks.

\begin{table}[t]
\centering
\scalebox{0.9}{
\begin{tabular}{|p{17cm}|}
\toprule

{\bf Advertising} is how a company \underline{\color{mynicegreen}{{\it cheats}} \color{black}{/} \color{mynicegreen}{grills}} people to buy their products, services or \underline{\color{mynicegreen}{{\it dreams}} \color{black}{/} \color{mynicegreen}{optics}} $\dots$ that draws \underline{\color{mynicegreen}{{\it silly}} \color{black}{/} \color{mynicegreen}{hotter}} attention towards these things. 
Companies use ads to try to get people to \underline{\color{mynicegreen}{{\it forget}} \color{black}{/} \color{mynicegreen}{fling}} their products, by showing them the good rather than the bad of their \underline{\color{mynicegreen}{{\it products}} \color{black}{/} \color{mynicegreen}{earrings}}.
For example, to make a \underline{\color{mynicegreen}{{\it burger}} \color{black}{/} \color{mynicegreen}{dynamite}} look tasty in advertising, it may be painted with brown food colors, sprayed with \underline{\color{mynicegreen}{{\it oil}} \color{black}{/} \color{mynicegreen}{crocodile}} to prevent it from going \underline{\color{mynicegreen}{{\it dull}} \color{black}{/} \color{mynicegreen}{graceful}}, and sesame seeds may be super-glued in place.
Advertising can bring new \underline{\color{mynicegreen}{{\it scapegoats}} \color{black}{/} \color{mynicegreen}{sickness}} and more sales for a business.
Advertising can be \underline{\color{mynicegreen}{{\it useless}} \color{black}{/} \color{mynicegreen}{grim}} $\dots$
\\
\midrule
{\bf Valentine's Day} is a \underline{\color{blue}{{\it fixture}} \color{black}{/} \color{blue}{prank}} that happens on February 14 $\dots$ when lovers show their \underline{\color{blue}{{\it toe}} \color{black}{/}\color{blue}{crush}} to each other $\dots$
by giving Valentine's cards or just a \underline{\color{blue}{{\it stinky}} \color{black}{/}\color{blue}{handy}} gift.
Some people \underline{\color{blue}{{\it kill}} \color{black}{/}\color{blue}{buffet}} one person and call them their Valentine as a gesture to show \underline{\color{blue}{{\it equipment}} \color{black}{/}\color{blue}{mischief}} and appreciation.
Valentine's Day is named for the \underline{\color{blue}{{\it gross}} \color{black}{/}\color{blue}{annoying}} Christian saint $\dots$ who performed \underline{\color{blue}{{\it gorillas}} \color{black}{/}\color{blue}{outdoors}} for couples who were not allowed to get married because their \underline{\color{blue}{{\it squids}} \color{black}{/}\color{blue}{circus}} did not agree with the connection
$\dots$
so the marriage was \underline{\color{blue}{{\it eaten}} \color{black}{/}\color{blue}{exploded}}.
Valentine gave the married couple flowers from his \underline{\color{blue}{{\it casket}} \color{black}{/}\color{blue}{problem}}.
That is why flowers play a very \underline{\color{blue}{{\it hungry}} \color{black}{/}\color{blue}{abusive}} role on Valentine's Day. \\
\bottomrule
\end{tabular}
}
\caption{Examples with YodaLib stories are rated as funnier than FT stories (top: \color{mynicegreen}{IN}\color{black}, down: \color{blue}{US}\color{black}. Filled-in word order: \underline{{\it FreeText}/{\sc YodaLib}}).
}
\label{tab:bestProposedHumorStories}
\end{table}

Table \ref{tab:bestProposedHumorStories} shows  example story snippets where YodaLib stories receive higher {\sc mfg} than FT stories.
Humor in the {\sc YodaLib} stories is largely via incongruity
(e.g., {\it advertisement to \underline{grill} people}, {\it Valentine's Day is a \underline{prank}}).
Yet, the {\sc YodaLib} approach is also able to generate small snippets of coherent and funny phrases:
Advertising is how a company \underline{grills} people to buy their products, it can bring \underline{sickness} and be \underline{grim};
Valentine's Day is a \underline{prank} to show \underline{mischief}, when lovers show their \underline{crush} to each other;
Batman is an \underline{abusive} superhero, a \underline{fiery} child, and grew up learning different ways to \underline{glare} (from Table \ref{tab:bestUSHumorStories}).

\begin{table*}[t]
\centering
\scalebox{0.9}{
\begin{tabular}{|p{17cm}|}
\toprule
{\bf Batman} is one of the most \underline{\color{blue}{{\it bland}} \color{black}{/} \color{blue}{abusive}} superheroes.
He was the second \underline{\color{blue}{{\it wimp}} \color{black}{/} \color{blue}{pest}} to be created
$\dots$ lives in the \underline{\color{blue}{{\it discombobulated}} \color{black}{/} \color{blue}{tool}} city of Gotham $\dots$ origin story as a \underline{\color{blue}{{\it moronic}} \color{black}{/} \color{blue}{fiery}} child, Bruce Wayne saw a robber \underline{\color{blue}{{\it kiss}} \color{black}{/} \color{blue}{spin}} his parents after the family left a \underline{\color{blue}{{\it bakery}} \color{black}{/} \color{blue}{teddy}} $\dots$ he did not want that kind of \underline{\color{blue}{{\it romance}} \color{black}{/} \color{blue}{wayne}} to happen to anyone else. He dedicated his life to \underline{\color{blue}{{\it terrorize}} \color{black}{/} \color{blue}{tolerate}} Gotham City. Wayne learned many different ways to \underline{\color{blue}{{\it grow}} \color{black}{/} \color{blue}{glare}} as he grew up $\dots$ \\
\bottomrule
\end{tabular}
}
\caption{An example story where US-FT humor is better than US-YodaLib humor
Filled-in word order: \underline{{\sc {\it FreeText}}/{\sc YodaLib}}.}
\label{tab:bestUSHumorStories}
\vspace{-0.15in}
\end{table*}

We note that similar to the observation in \cite{Hossain2017}, when humans write funnier stories, they do so by incorporating more coherence.
Table \ref{tab:bestUSHumorStories} shows a US-written {\sc FT} story with high {\sc mfg} from US judges (IN judges often give higher grades to {\sc YodaLib} stories).
The US turker portrays Batman as a bland and an uninteresting person who is weak, cowardly and moronic, living in a disconcerted city of Gotham.
Humor is generated by portraying the title concepts consistently with surprising and unexpected views. 
 It is interesting that Batman is depicted with two traits consistently: his bland and moronic personality, and his opposition to romance in Gotham due to a robber kissing his parents.
This is striking, as it illustrates how good skilled humans of IN and US origin can be at generating humor via multiple coherent and meaningful concepts.
However, this is seen in a very few stories, possibly due to the biases humans may have in what they consider humorous. 
Nevertheless, it provides us with future directions to pursue to generate demographic-aware humor of even higher quality in terms of coherence.


\section{Conclusion}
\label{sec:conclusions}
In this paper, we studied location-specific humor generation in Mad Libs stories.
We first collected a novel location-specific humor dataset on AMT, by selecting players and judges to obtain ground truth data.
Next, we proposed an automated location-specific humor generation framework to generate possible candidates to fill-in the blanks,  to rank these candidates based on their humor contributions to form funny sentences, and to complete the stories by selecting the best transformations for the constituting sentences.
Our approach outperformed a simple language model and human players (in most cases) in generating funny stories.

We also performed a detailed demographic-based analysis of our dataset.
We found that humor created with US slant is in general preferred to IN slant humor by both IN and US judges.
IN judges seemed to have a more lenient outlook towards humor, while US judges have higher expectations possibly in terms of coherence, which is also reflected in the better quality humor generated in the US-specific setting.
When turkers wrote funnier stories, they did so in a coherent manner,
indicating the vast potential of coherence in generating humor, contrary to the general incongruent take of Mad Libs.
Humor from our approach is generated primarily via incongruity and deviation, despite our preliminary measures to incorporate coherence.
We believe that our proposed BERT-based approach is general, and can be used for other affect-based NLP tasks with minor changes.
In the future, we aim to extend our research in several directions, using (1) a larger number of and a wider variety of Mad Lib-like stories; and (2) other demographic dimensions (e.g., gender, age group, occupation), and more groups within each dimension (e.g., Singapore, Canada, England for location; Arts, Engineering, Fashion, Publishing for occupation).
We also plan to take steps toward understanding what makes textual humor coherent, and go beyond our word and sentence similarity measures to generate more coherent and funny stories.

\bibliographystyle{coling}
\bibliography{coling2020}

\begin{thebibliography}{}

\bibitem[\protect\citename{Alden \bgroup et al.\egroup }1993]{Alden1993}
Dana~L Alden, Wayne~D Hoyer, and Chol Lee.
\newblock 1993.
\newblock Identifying global and culture-specific dimensions of humor in
  advertising: A multinational analysis.
\newblock {\em The Journal of Marketing}, pages 64--75.

\bibitem[\protect\citename{Apte}1985]{Apte1985}
Mahadev~L Apte.
\newblock 1985.
\newblock {\em Humor and laughter: An anthropological approach}.
\newblock Cornell Univ Pr.

\bibitem[\protect\citename{Attardo and Raskin}1991]{Attardo1991}
Salvatore Attardo and Victor Raskin.
\newblock 1991.
\newblock Script theory revis (it) ed: Joke similarity and joke representation
  model.
\newblock {\em Humor-International Journal of Humor Research}, 4(3-4):293--348.

\bibitem[\protect\citename{Attardo}2010]{Attardo2010}
Salvatore Attardo.
\newblock 2010.
\newblock {\em Linguistic theories of humor}, volume~1.
\newblock Walter de Gruyter.

\bibitem[\protect\citename{Bamman \bgroup et al.\egroup }2014]{Bamman2014}
David Bamman, Chris Dyer, and Noah~A. Smith.
\newblock 2014.
\newblock {Distributed representations of geographically situated language}.
\newblock In {\em Proceedings of the 52nd Annual Meeting of the Association for
  Computational Linguistics (Volume 2: Short Papers) (ACL 2014)}, pages
  828--834.

\bibitem[\protect\citename{Bertero and Fung}2016]{Bertero2016}
Dario Bertero and Pascale Fung.
\newblock 2016.
\newblock A long short-term memory framework for predicting humor in dialogues.
\newblock In {\em Proceedings of the 2016 Conference of the North American
  Chapter of the Association for Computational Linguistics: Human Language
  Technologies}, pages 130--135.

\bibitem[\protect\citename{Binsted \bgroup et al.\egroup }1997]{Binsted1997}
Kim Binsted, Helen Pain, and Graeme~D Ritchie.
\newblock 1997.
\newblock Children's evaluation of computer-generated punning riddles.
\newblock {\em Pragmatics \& Cognition}, 5(2):305--354.

\bibitem[\protect\citename{Blinov \bgroup et al.\egroup }2019]{Blinov2019}
Vladislav Blinov, Valeria Bolotova-Baranova, and Pavel Braslavski.
\newblock 2019.
\newblock Large dataset and language model fun-tuning for humor recognition.
\newblock In {\em Proceedings of the 57th Annual Meeting of the Association for
  Computational Linguistics}, pages 4027--4032, Florence, Italy, July.
  Association for Computational Linguistics.

\bibitem[\protect\citename{Devlin \bgroup et al.\egroup }2019]{Devlin2019}
Jacob Devlin, Ming-Wei Chang, Kenton Lee, and Kristina Toutanova.
\newblock 2019.
\newblock {BERT}: Pre-training of deep bidirectional transformers for language
  understanding.
\newblock In {\em Proceedings of the 2019 Conference of the North {A}merican
  Chapter of the Association for Computational Linguistics: Human Language
  Technologies, Volume 1 (Long and Short Papers)}, pages 4171--4186,
  Minneapolis, Minnesota, June. Association for Computational Linguistics.

\bibitem[\protect\citename{Duncan \bgroup et al.\egroup }1990]{Duncan1990}
W~Jack Duncan, Larry~R Smeltzer, and Terry~L Leap.
\newblock 1990.
\newblock Humor and work: Applications of joking behavior to management.
\newblock {\em Journal of Management}, 16(2):255--278.

\bibitem[\protect\citename{Eckert and McConnell-Ginet}2013]{Eckert2013}
Penelope Eckert and Sally McConnell-Ginet.
\newblock 2013.
\newblock {\em Language and gender}.
\newblock Cambridge University Press.

\bibitem[\protect\citename{Garimella and Mihalcea}2016]{Garimella2016b}
Aparna Garimella and Rada Mihalcea.
\newblock 2016.
\newblock Zooming in on gender differences in social media.
\newblock {\em PEOPLES 2016}, page~1.

\bibitem[\protect\citename{Garimella \bgroup et al.\egroup
  }2016]{Garimella2016}
Aparna Garimella, Rada Mihalcea, and James Pennebaker.
\newblock 2016.
\newblock Identifying cross-cultural differences in word usage.
\newblock In {\em Proceedings of the 26th International Conference on
  Computational Linguistics: Technical Papers (COLING 2016)}, pages 674--683,
  Osaka, Japan.

\bibitem[\protect\citename{Garimella \bgroup et al.\egroup
  }2017]{Garimella2017}
Aparna Garimella, Carmen Banea, and Rada Mihalcea.
\newblock 2017.
\newblock Demographic-aware word associations.
\newblock In {\em Proceedings of the 2017 Conference on Empirical Methods in
  Natural Language Processing}, pages 2275--2285.

\bibitem[\protect\citename{Goodman}1992]{Goodman1992}
Lizbeth Goodman.
\newblock 1992.
\newblock {\em Gender and humour}.
\newblock na.

\bibitem[\protect\citename{Hasan \bgroup et al.\egroup }2019]{Hasan2019}
Md~Kamrul Hasan, Wasifur Rahman, Amir Zadeh, Jianyuan Zhong, Md~Iftekhar
  Tanveer, Louis-Philippe Morency, et~al.
\newblock 2019.
\newblock Ur-funny: A multimodal language dataset for understanding humor.
\newblock {\em arXiv preprint arXiv:1904.06618}.

\bibitem[\protect\citename{Hay}2000]{Hay2000}
Jennifer Hay.
\newblock 2000.
\newblock Functions of humor in the conversations of men and women.
\newblock {\em Journal of pragmatics}, 32(6):709--742.

\bibitem[\protect\citename{Hossain \bgroup et al.\egroup }2017]{Hossain2017}
Nabil Hossain, John Krumm, Lucy Vanderwende, Eric Horvitz, and Henry Kautz.
\newblock 2017.
\newblock Filling the blanks (hint: plural noun) for mad libs humor.
\newblock In {\em Proceedings of the 2017 Conference on Empirical Methods in
  Natural Language Processing}, pages 638--647.

\bibitem[\protect\citename{Hossain \bgroup et al.\egroup }2019]{Hossain2019}
Nabil Hossain, John Krumm, and Michael Gamon.
\newblock 2019.
\newblock {``}{P}resident vows to cut {\textless}taxes{\textgreater} hair{''}:
  Dataset and analysis of creative text editing for humorous headlines.
\newblock In {\em Proceedings of the 2019 Conference of the North {A}merican
  Chapter of the Association for Computational Linguistics: Human Language
  Technologies, Volume 1 (Long and Short Papers)}, pages 133--142, Minneapolis,
  Minnesota, June. Association for Computational Linguistics.

\bibitem[\protect\citename{Hovy}2015]{Hovy2015}
Dirk Hovy.
\newblock 2015.
\newblock Demographic factors improve classification performance.
\newblock In {\em Proceedings of the 53rd Annual Meeting of the Association for
  Computational Linguistics and the 7th International Joint Conference on
  Natural Language Processing (ACL 2015)}, pages 752--762, Beijing, China.

\bibitem[\protect\citename{Johnstone}2010]{Johnstone2010}
Barbara Johnstone.
\newblock 2010.
\newblock Language and place.

\bibitem[\protect\citename{Kiddon and Brun}2011]{Kiddon2011}
Chloe Kiddon and Yuriy Brun.
\newblock 2011.
\newblock That's what she said: double entendre identification.
\newblock In {\em Proceedings of the 49th Annual Meeting of the Association for
  Computational Linguistics: Human Language Technologies: short papers-Volume
  2}, pages 89--94. Association for Computational Linguistics.

\bibitem[\protect\citename{Kramarae}1981]{Kramarae1981}
Cheris Kramarae.
\newblock 1981.
\newblock Women and men speaking: Frameworks for analysis.

\bibitem[\protect\citename{Krippendorff}1970]{Krippendorff1970}
Klaus Krippendorff.
\newblock 1970.
\newblock Estimating the reliability, systematic error and random error of
  interval data.
\newblock {\em Educational and Psychological Measurement}, 30(1):61--70.

\bibitem[\protect\citename{Lin \bgroup et al.\egroup }2018]{Lin2018}
Bill~Yuchen Lin, Frank~F. Xu, Kenny Zhu, and Seung-won Hwang.
\newblock 2018.
\newblock Mining cross-cultural differences and similarities in social media.
\newblock In {\em Proceedings of the 56th Annual Meeting of the Association for
  Computational Linguistics (Volume 1: Long Papers)}, pages 709--719,
  Melbourne, Australia, July. Association for Computational Linguistics.

\bibitem[\protect\citename{Loveys \bgroup et al.\egroup }2018]{Loveys2018}
Kate Loveys, Jonathan Torrez, Alex Fine, Glen Moriarty, and Glen Coppersmith.
\newblock 2018.
\newblock Cross-cultural differences in language markers of depression online.
\newblock In {\em Proceedings of the Fifth Workshop on Computational
  Linguistics and Clinical Psychology: From Keyboard to Clinic}, pages 78--87,
  New Orleans, LA, June. Association for Computational Linguistics.

\bibitem[\protect\citename{Mihalcea and Strapparava}2006]{Mihalcea2006b}
Rada Mihalcea and Carlo Strapparava.
\newblock 2006.
\newblock Learning to laugh (automatically): Computational models for humor
  recognition.
\newblock {\em Computational Intelligence}, 22(2):126--142.

\bibitem[\protect\citename{Morreall}2012]{Morreall2012}
John Morreall.
\newblock 2012.
\newblock Philosophy of humor.

\bibitem[\protect\citename{Mostafazadeh \bgroup et al.\egroup
  }2017]{Mostafazadeh2017}
Nasrin Mostafazadeh, Michael Roth, Annie Louis, Nathanael Chambers, and James
  Allen.
\newblock 2017.
\newblock Lsdsem 2017 shared task: The story cloze test.
\newblock In {\em Proceedings of the 2nd Workshop on Linking Models of Lexical,
  Sentential and Discourse-level Semantics}, pages 46--51.

\bibitem[\protect\citename{O'Shannon}2012]{O2012}
Dan O'Shannon.
\newblock 2012.
\newblock {\em What are You Laughing At?: A Comprehensive Guide to the Comedic
  Event}.
\newblock A\&C Black.

\bibitem[\protect\citename{Petrovi{\'c} and Matthews}2013]{Petrovic2013}
Sa{\v{s}}a Petrovi{\'c} and David Matthews.
\newblock 2013.
\newblock Unsupervised joke generation from big data.
\newblock In {\em Proceedings of the 51st Annual Meeting of the Association for
  Computational Linguistics (Volume 2: Short Papers)}, volume~2, pages
  228--232.

\bibitem[\protect\citename{Price and Stern}1974]{price1974}
Roger Price and Leonard Stern.
\newblock 1974.
\newblock {\em Mad Libs: World's Greatest Party Game: a Do-it-yourself Laugh
  Kit}.
\newblock Number~1. Mad Libs.

\bibitem[\protect\citename{Raz}2012]{Raz2012}
Yishay Raz.
\newblock 2012.
\newblock Automatic humor classification on twitter.
\newblock In {\em Proceedings of the 2012 Conference of the North American
  Chapter of the Association for Computational Linguistics: Human Language
  Technologies: Student Research Workshop}, pages 66--70. Association for
  Computational Linguistics.

\bibitem[\protect\citename{Robinson and Smith-Lovin}2001]{Robinson2001}
Dawn~T Robinson and Lynn Smith-Lovin.
\newblock 2001.
\newblock Getting a laugh: Gender, status, and humor in task discussions.
\newblock {\em Social Forces}, 80(1):123--158.

\bibitem[\protect\citename{Stock and Strapparava}2002]{Stock2002}
Oliviero Stock and Carlo Strapparava.
\newblock 2002.
\newblock Hahacronym: Humorous agents for humorous acronyms.
\newblock {\em Stock, Oliviero, Carlo Strapparava, and Anton Nijholt. Eds},
  pages 125--135.

\bibitem[\protect\citename{Tresselt and Mayzner}1964]{Tresselt64}
Margaret~E. Tresselt and Mark~S. Mayzner.
\newblock 1964.
\newblock The {Kent-Rosanoff} word association: Word association norms as a
  function of age.
\newblock {\em Psychonomic Science}, 1(1-12):65--66.

\bibitem[\protect\citename{Vaswani \bgroup et al.\egroup }2017]{Vaswani2017}
Ashish Vaswani, Noam Shazeer, Niki Parmar, Jakob Uszkoreit, Llion Jones,
  Aidan~N Gomez, {\L}ukasz Kaiser, and Illia Polosukhin.
\newblock 2017.
\newblock Attention is all you need.
\newblock In {\em Advances in neural information processing systems}, pages
  5998--6008.

\bibitem[\protect\citename{Volkova \bgroup et al.\egroup }2013]{Volkova2013}
Svitlana Volkova, Theresa Wilson, and David Yarowsky.
\newblock 2013.
\newblock Exploring demographic language variations to improve multilingual
  sentiment analysis in social media.
\newblock In {\em Proceedings of the 2013 Conference on Empirical Methods in
  Natural Language Processing (EMNLP 2013)}, number October, pages 1815--1827,
  Seattle, WA, USA.

\bibitem[\protect\citename{Weems}2014]{Weems2014}
Scott Weems.
\newblock 2014.
\newblock {\em Ha!: The science of when we laugh and why}.
\newblock Basic Books (AZ).

\bibitem[\protect\citename{Wilkins and Eisenbraun}2009]{Wilkins2009}
Julia Wilkins and Amy~Janel Eisenbraun.
\newblock 2009.
\newblock Humor theories and the physiological benefits of laughter.
\newblock {\em Holistic nursing practice}, 23(6):349--354.

\bibitem[\protect\citename{Wilson \bgroup et al.\egroup }2016]{Wilson2016}
Steven Wilson, Rada Mihalcea, Ryan Boyd, and James Pennebaker.
\newblock 2016.
\newblock Disentangling topic models: A cross-cultural analysis of personal
  values through words.
\newblock In {\em Proceedings of the First Workshop on {NLP} and Computational
  Social Science}, pages 143--152, Austin, Texas, November. Association for
  Computational Linguistics.

\bibitem[\protect\citename{Zhang and Liu}2014]{Zhang2014}
Renxian Zhang and Naishi Liu.
\newblock 2014.
\newblock Recognizing humor on twitter.
\newblock In {\em Proceedings of the 23rd ACM International Conference on
  Conference on Information and Knowledge Management}, pages 889--898. ACM.

\bibitem[\protect\citename{Zhu \bgroup et al.\egroup }2015]{Zhu2015}
Yukun Zhu, Ryan Kiros, Rich Zemel, Ruslan Salakhutdinov, Raquel Urtasun,
  Antonio Torralba, and Sanja Fidler.
\newblock 2015.
\newblock Aligning books and movies: Towards story-like visual explanations by
  watching movies and reading books.
\newblock In {\em Proceedings of the IEEE international conference on computer
  vision}, pages 19--27.

\end{thebibliography}

\end{document}


\maketitle

\appendix
\section{Appendices}

\subsection{Mechanical Turk Details}
\label{sec:MechanicalTurkDetails}
Each task for judge selection are initially published for 60 turker responses, and player selection for 30 responses.
However, due to fewer qualified IN judges and players hence obtained, we republished these tasks for 40 more IN turkers for judge selection and 20 more IN turkers for player selection, to obtain comparable number of judges and players from IN as from US.
\begin{figure*}[th]
    \centering
    \includegraphics[width=1.0\linewidth, height=4.5cm]{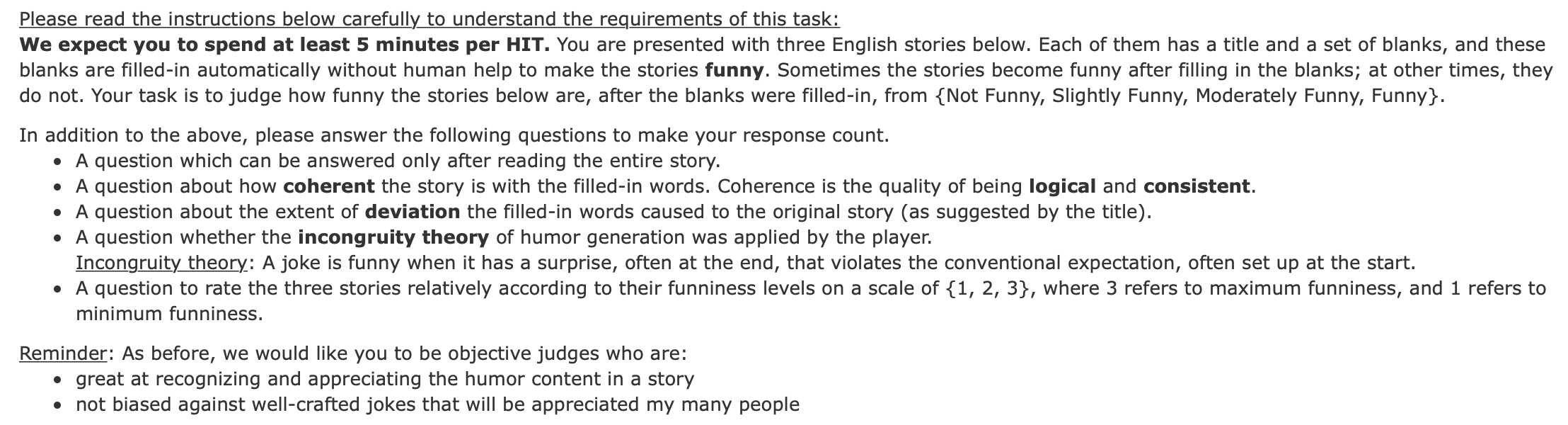}
    \caption{Judge Instructions from AMT.}
    \label{fig:judgeInstructions}
\end{figure*}

\begin{figure*}[th]
    \centering
    \includegraphics[width=1.0\linewidth, height=6.5cm]{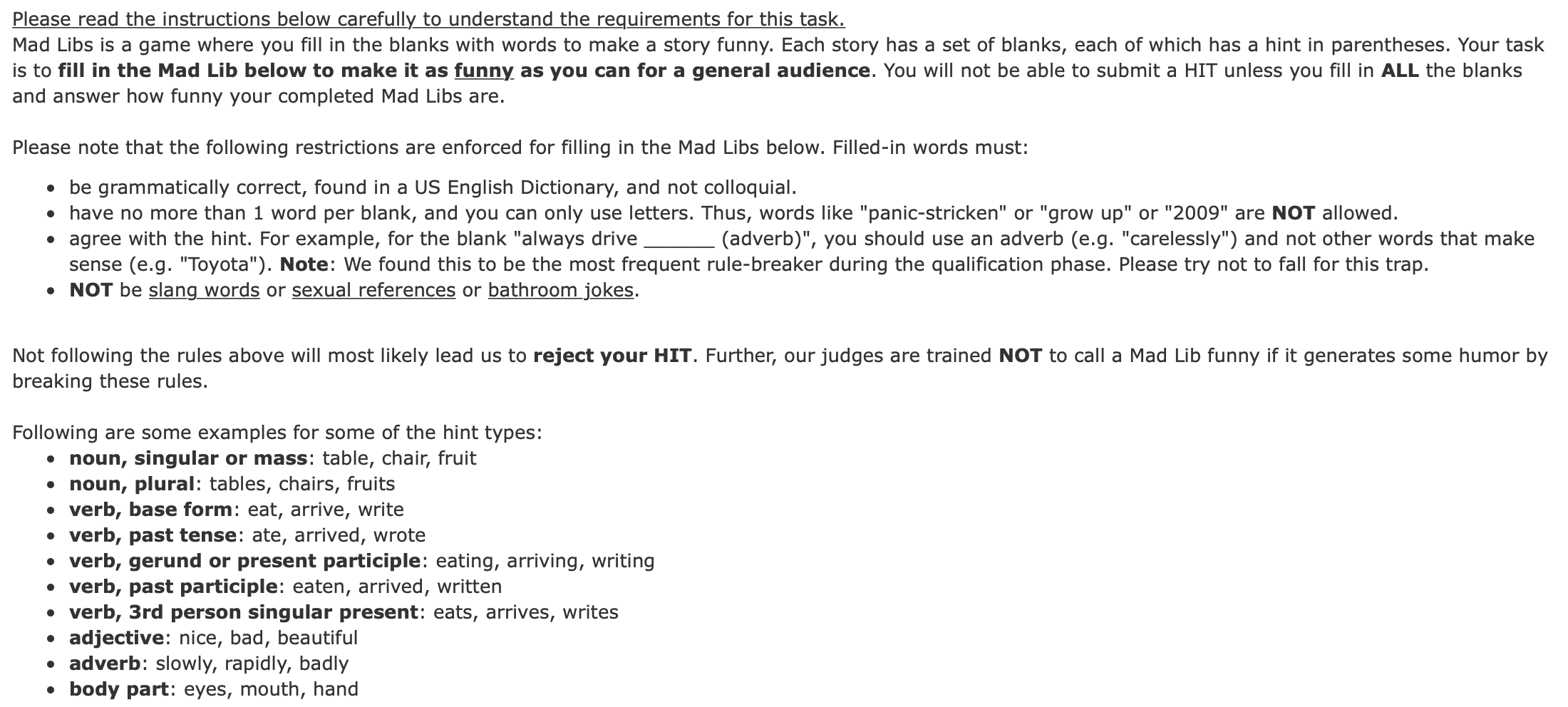}
    \caption{Player Instructions from AMT.}
    \label{fig:playerInstructions}
\end{figure*}
 
Figures \ref{fig:judgeInstructions} and \ref{fig:playerInstructions} show screenshots of the instructions provided to the turkers on AMT from each country, during the judge selection and player selection phases.
The players are paid \$0.50 in the player selection and annotation phases, while the judges receive \$0.25, \$0.10 and \$0.50 in the judge selection, player selection and annotation phases, respectively.
The total cost is approximately \$1,200.

\subsection{Wikipedia Sentences Details}
Sentences with word count $\mathrm{M}\pm5$ are sampled from English Wikipedia; M: Median of word counts of train sentences.

\subsection{Example Stories}
\begin{table*}[th]
\centering
\scalebox{0.9}{
\begin{tabular}{p{16cm}}
\toprule
{\bf Scuba Diving} is a sport where people can swim under \underline{\color{mynicegreen}{{{\it beer}}} \color{black}{/} \color{mynicegreen}{{feather}}} for a long time, using a tank filled with compressed \underline{\color{mynicegreen}{{\textit{nitrogen}}} \color{black}{/} \color{mynicegreen}{cigarette}}. The tank is a \underline{\color{mynicegreen}{{\it hollow}} \color{black}{/} \color{mynicegreen}{dusty}} cylinder made of steel or \underline{\color{mynicegreen}{{\it rubber}} \color{black}{/} \color{mynicegreen}{star}}. A scuba diver \underline{\color{mynicegreen}{{\it walks}} \color{black}{/} \color{mynicegreen}{burns}} underwater by using fins attached to the \underline{\color{mynicegreen}{{\it legs}} \color{black}{/} \color{mynicegreen}{windows}}. They also use \underline{\color{mynicegreen}{{\it coconut}} \color{black}{/} \color{mynicegreen}{hospitality}} such as a dive mask to \underline{\color{mynicegreen}{{\it hide}} \color{black}{/} \color{mynicegreen}{burn}} underwater vision and equipment to control \underline{\color{mynicegreen}{{\it air}} \color{black}{/} \color{mynicegreen}{crouch}}. A person must take a \underline{\color{mynicegreen}{{\it course}} \color{black}{/} \color{mynicegreen}{cigarette}} class before going scuba diving. This proves that they have been trained on how to \underline{\color{mynicegreen}{{\it bunk}} \color{black}{/} \color{mynicegreen}{punch}} the equipment and dive \underline{\color{mynicegreen}{{\it deep}} \color{black}{/} \color{mynicegreen}{cold}}. Some tourist attractions have a \underline{\color{mynicegreen}{{\it simple}} \color{black}{/} \color{mynicegreen}{deadly}} course on certification and then the instructors \underline{\color{mynicegreen}{{\it bunk}} \color{black}{/} \color{mynicegreen}{sing}} the class in a \underline{\color{mynicegreen}{{\it simple}} \color{black}{/} \color{mynicegreen}{textile}} dive, all in one day. \\
\midrule
{\bf Cats} are the most \underline{\color{blue}{{\it naughty}} \color{black}{/} \color{blue}foolish} pets in the world. They were probably first kept because they ate \underline{\color{blue}{{\it dogs}} \color{black}{/} \color{blue}lawyers}. Later cats were \underline{\color{blue}{{\it praised}} \color{black}{/} \color{blue}disgusted} because they are \underline{\color{blue}{{\it ferocious}} \color{black}{/} \color{blue}psychotic} and they are good \underline{\color{blue}{{\it thinkers}} \color{black}{/} \color{blue}radicals}. Cats are active carnivores, meaning they hunt \underline{\color{blue}{{\it small}} \color{black}{/} \color{blue}wicked} prey. They mainly prey on small mammals, like \underline{\color{blue}{{\it dogs}} \color{black}{/} \color{blue}dice}. Their main method of \underline{\color{blue}{{\it hunting}} \color{black}{/} \color{blue}baking} is stalk and \underline{\color{blue}{{\it kiss}} \color{black}{/} \color{blue}kid}. While dogs have great \underline{\color{blue}{{\it allergy}} \color{black}{/} \color{blue}gore} and they will catch prey over long distances, cats are extremely \underline{\color{blue}{{\it romantic}} \color{black}{/} \color{blue}creepy}, but only over short distances. The cat creeps towards a chosen victim, keeping its \underline{\color{blue}{{\it mouth}} \color{black}{/} \color{blue}buster} flat and near to the \underline{\color{blue}{{\it dog}} \color{black}{/} \color{blue}booth} so that it cannot be \underline{\color{blue}{{\it sensed}} \color{black}{/} \color{blue}cashed} easily, until it is close enough for a rapid \underline{\color{blue}{{\it kiss}} \color{black}{/} \color{blue}neutron} or pounce.
\\
\bottomrule
\end{tabular}
}
\caption{Example stories with highest improvements of {\sc YodaLib} over {\sc FreeText} with filled-in words in the following order: \underline{\color{mynicegreen}{\sc {\it FreeText}} \color{black}{/} \color{mynicegreen}{YodaLib}} for Indian humor ({\it Scuba Diving}), and \underline{\color{blue}{\sc {\it FreeText}} \color{black}{/} \color{blue}{YodaLib}} for US humor ({\it Cats}).}
\label{tab:bestHumorProposedOverFreeText}
\end{table*}
Table \ref{tab:bestHumorProposedOverFreeText} shows two stories, one of Indian humor and the other of US humor, for which the {\sc YodaLib} outperforms the {\sc FT} by a large margin when graded by IN judges.
As expected, incongruity is significant in generating humor in the stories via the {\sc YodaLib} approach ({\it e.g.,} {\it tank filled with \underline{nitrogen}} vs {\it tank filled with \underline{cigarette}} in ``Scuba Diving," {\it {\it they ate \underline{dogs}} vs {\it they are \underline{lawyers}}} in ``Cats"), while the {\sc FT} stories, though being less humorous, are slightly more coherent ({\it e.g.,} {\it scuba diver \underline{walks} underwater using fins attached to \underline{legs}} in ``Scuba Diving," {\it main method of \underline{hunting} is stalk and \underline{kiss} ... cats are extremely \underline{romantic}} in ``Cats").

\begin{table*}[th]
\centering
\scalebox{0.9}{
\begin{tabular}{p{16cm}}
\toprule
{\bf Valentine's Day} is a \underline{\color{mynicegreen}{{\it training}} \color{black}{/} \color{mynicegreen}{pigeon}} that happens on February 14. It is the day of the year when lovers show their \underline{\color{mynicegreen}{{\it drama}} \color{black}{/} \color{mynicegreen}{bird}} to each other. This can be done by giving \underline{\color{mynicegreen}{{\it sticks}} \color{black}{/} \color{mynicegreen}{sports}}, flowers, Valentine's cards or just a/an \underline{\color{mynicegreen}{{\it bad}} \color{black}{/} \color{mynicegreen}{ultraviolet}} gift. Some people \underline{\color{mynicegreen}{{\it kill}} \color{black}{/} \color{mynicegreen}{drink}} one person and call them their ``Valentine" as a gesture to show \underline{\color{mynicegreen}{{\it love}} \color{black}{/} \color{mynicegreen}{nero}} and appreciation. Valentine's Day is named for the \underline{\color{mynicegreen}{{\it mad}} \color{black}{/} \color{mynicegreen}{kind}} Christian saint named Valentine. He was a bishop who performed \underline{\color{mynicegreen}{{\it dramas}} \color{black}{/} \color{mynicegreen}{strings}} for couples who were not allowed to get married because their \underline{\color{mynicegreen}{{\it friends}} \color{black}{/} \color{mynicegreen}{goddesses}} did not agree with the connection or because the bridegroom was a soldier or a/an \underline{\color{mynicegreen}{{\it joker}} \color{black}{/} \color{mynicegreen}{eel}}, so the marriage was \underline{\color{mynicegreen}{{\it solved}} \color{black}{/} \color{mynicegreen}{chickened}}. Valentine gave the married couple flowers from his \underline{\color{mynicegreen}{{\it kitchen}} \color{black}{/} \color{mynicegreen}{name}}. That is why flowers play a very \underline{\color{mynicegreen}{{\it tasty}} \color{black}{/} \color{mynicegreen}{mythological}} role on Valentine's Day. This did not \underline{\color{mynicegreen}{{\it attract}} \color{black}{/} \color{mynicegreen}{bat}} the emperor, and Valentine was \underline{\color{mynicegreen}{{\it praised}} \color{black}{/} \color{mynicegreen}{melted}} because of his Christian \underline{\color{mynicegreen}{{\it drama}} \color{black}{/} \color{mynicegreen}{bath}}.
\\
\bottomrule
\end{tabular}
}
\caption{The best {\it Valentine's Day} story obtained for IN humor using {\sc FT} approach. Filled-in words in the following order: \underline{\color{mynicegreen}{\sc {\it FT}} \color{black}{/} \color{mynicegreen}{YodaLib}}.}
\label{tab:bestIndianHumorStoriesAppendix}
\end{table*}
Tables \ref{tab:bestIndianHumorStoriesAppendix} and \ref{tab:bestUSHumorStoriesAppendix} show IN and US stories from {\sc FT} approach that received the highest {\sc mfg} by US judges.
US turkers describe Snoring as a disturbing and murderous noise that is symbolic of stupidity and lack of valor, and Scuba Diving as a violent sport involving deadly dives and fatal courses that takes place under lava, and intended to destroy the equipment and avoid panicking while doing so.
Humor in these stories is generated by portraying the title concepts as notions that are surprising and unexpected, and doing so consistently. 

Much of the humor generated via the {\sc YodaLib} approach is through incongruity -- the filled-in words are funny as they do not match the expectation of the reader ({\it e.g.,} {\it Snoring is the \underline{robot} that people make when they are \underline{brooding}}, {\it a scuba diver \underline{steals} underwater using fins attached to the \underline{bombs}}, {\it Batman is an \underline{abusive} superhero}).
Occasionally, the {\sc YodaLib} approach is able to generate small snippets of coherent and funny phrases.
For example, the tourist attractions of Scuba Diving are described as having \underline{risky} courses where the instructors \underline{attack} the class; Batman is described as an \underline{abusive} superhero who was \underline{fiery} as a child, and grew up learning different ways to \underline{glare}; Valentine's Day is named after a \underline{kind} bishop, who performed \underline{strings} for couples whose marriages were not allowed due to the disagreement of their \underline{goddesses}, and the flowers given by him play a \underline{mythological} role.

\begin{table*}[t]
\centering
\scalebox{0.9}{
\begin{tabular}{p{16cm}}
\toprule
{\bf Snoring} is the \underline{\color{blue}{{\it disturbance}} \color{black}{/} \color{blue}{robot}} that people often make when they are \underline{\color{blue}{\textit{relaxing}} \color{black}{/} \color{blue}{brooding}}. It is often caused by a blocked \underline{\color{blue}{\textit{intestine}} \color{black}{/} \color{blue}{laugh}} or throat. The noise is often \underline{\color{blue}{{\it murderous}} \color{black}{/} \color{blue}{societal}}, as it is made by \underline{\color{blue}{{\it air}} \color{black}{/} \color{blue}{madman}} passing through the \underline{\color{blue}{{\it nasal}} \color{black}{/} \color{blue}{presidential}} passages or the \underline{\color{blue}{{\it mouth}} \color{black}{/} \color{blue}{runway}}. Research suggests that snoring is one of the factors of \underline{\color{blue}{{\it job}} \color{black}{/}  \color{blue}{fame}} deprivation. It also causes daytime \underline{\color{blue}{{\it stupidity}} \color{black}{/} \color{blue}{bum}}, irritability, and lack of \underline{\color{blue}{{\it valor}} \color{black}{/} \color{blue}{buck}}. Snoring can cause significant \underline{\color{blue}{{\it moral}} \color{black}{/} \color{blue}{naval}} and social damage to \underline{\color{blue}{{\it sociopaths}} \color{black}{/} \color{blue}{wolves}}. So far, there is no certain \underline{\color{blue}{{\it milkshake}} \color{black}{/} \color{blue}{}} available that can \underline{\color{blue}{{\it completely}} \color{black}{/} \color{blue}{bunny}} stop snoring.
 \\
\midrule
{\bf Scuba Diving} is a sport where people can swim under \underline{\color{blue}{{\it lava}} \color{black}{/} \color{blue}{{genie}}} for a long time, using a tank filled with compressed \underline{\color{blue}{{\it dough}} \color{black}{/} \color{blue}{brute}}. The tank is a \underline{\color{blue}{{\it rusty}} \color{black}{/} \color{blue}{stray}} cylinder made of steel or \underline{\color{blue}{{\it coconut}} \color{black}{/} \color{blue}{sailor}}. A scuba diver \underline{\color{blue}{{\it sprints}} \color{black}{/} \color{blue}{steals}} underwater by using fins attached to the \underline{\color{blue}{{\it ears}} \color{black}{/} \color{blue}{bombs}}. They also use \underline{\color{blue}{{\it camouflage}} \color{black}{/} \color{blue}{mister}} such as a dive mask to \underline{\color{blue}{{\it inhibit}} \color{black}{/} \color{blue}{pardon}} underwater vision and equipment to control \underline{\color{blue}{{\it panicking}} \color{black}{/} \color{blue}{bombing}}. A person must take a \underline{\color{blue}{{\it skydiving}} \color{black}{/} \color{blue}{crack}} class before going scuba diving. This proves that they have been trained on how to \underline{\color{blue}{{\it destroy}} \color{black}{/} \color{blue}{sniff}} the equipment and dive \underline{\color{blue}{{\it violently}} \color{black}{/} \color{blue}{willingly}}. Some tourist attractions have a \underline{\color{blue}{{\it fatal}} \color{black}{/} \color{blue}{risky}} course on certification and then the instructors \underline{\color{blue}{{\it leave}} \color{black}{/} \color{blue}{attack}} the class in a \underline{\color{blue}{{\it deadly}} \color{black}{/} \color{blue}{curly}} dive, all in one day. \\
\midrule
{\bf Batman} is a fictional character and one of the most \underline{\color{blue}{{\it bland}} \color{black}{/} \color{blue}{abusive}} superheroes. He was the second \underline{\color{blue}{{\it wimp}} \color{black}{/} \color{blue}{pest}} to be created, after Superman. Batman began in comic books and he was later \underline{\color{blue}{{\it exposed}} \color{black}{/} \color{blue}{fired}} in several movies, TV programs, and books. Batman lives in the \underline{\color{blue}{{\it discombobulated}} \color{black}{/} \color{blue}{tool}} city of Gotham. When he is not in \underline{\color{blue}{{\it costume}} \color{black}{/} \color{blue}{wink}}, he is Bruce Wayne, a very \underline{\color{blue}{{\it dense}} \color{black}{/} \color{blue}{pacific}} businessman. Batman's origin story is that as a \underline{\color{blue}{{\it moronic}} \color{black}{/} \color{blue}{fiery}} child, Bruce Wayne saw a robber \underline{\color{blue}{{\it kiss}} \color{black}{/} \color{blue}{spin}} his parents after the family left a \underline{\color{blue}{{\it bakery}} \color{black}{/} \color{blue}{teddy}}. Bruce decided that he did not want that kind of \underline{\color{blue}{{\it romance}} \color{black}{/} \color{blue}{wayne}} to happen to anyone else. He dedicated his life to \underline{\color{blue}{{\it terrorize}} \color{black}{/} \color{blue}{tolerate}} Gotham City. Wayne learned many different ways to \underline{\color{blue}{{\it grow}} \color{black}{/} \color{blue}{glare}} as he grew up. As an adult, he wore a \underline{\color{blue}{{\it diaper}} \color{black}{/} \color{blue}{bird}} to protect his \underline{\color{blue}{{\it skin}} \color{black}{/} \color{blue}{drill}} while fighting \underline{\color{blue}{{\it nuns}} \color{black}{/} \color{blue}{pneumonia}} in Gotham. \\
\bottomrule
\end{tabular}
}
\caption{The best stories for {\it Snoring}, {\it Scuba Diving} and {\it Batman} obtained for US humor. Filled-in words in the following order: \underline{\color{blue}{\sc {\it FT}} \color{black}{/} \color{blue}{YodaLib}}.}
\label{tab:bestUSHumorStoriesAppendix}
\end{table*}
Also, some of the culture-specific concepts surface in the stories generated via the {\sc YodaLib} approach.
Some examples for Indian stories include Valentine's Day with reference to {\it gods}, {\it goddesses} and {\it mythology}, Batman fighting {\it bombay} in Gotham, etc., while for US stories, a Beauty Contest is described in terms of {\it yankee} contest, Batman fights {\it Roosevelt} in Gotham and so on.
The relationship between culture-specific word occurrences in the generated stories and the culture of the target audience is to be explored in detail in the future.